# Dynamic Speed Guidance for CAV Ramp Merging in Non-Cooperative Environment: An On-Site Experiment

Wei Ji[1], Yechi Ma[1], Guangzhang Cui[1], Xiaotian Qin[2], Wei Hua[1*]

[1] *Research Center for Intelligent Transportation, Zhejiang Lab*
*(email: {jiw, mayechi, cuiguangzhang, huawei}@zhejianglab.com)*

[2] *School of Intelligent Systems Engineering, Sun Yat-sen University, China*
*(email: qinxt3@mail2.sysu.edu.cn)*

**Abstract**: Ramp merging is a typical application of cooperative intelligent transportation system (C-ITS). Vehicle trajectories perceived by roadside sensors are importation complement to the limited visual field of on-board perception. Vehicle tracking and trajectory denoising algorithm is proposed in this paper to take full advantage of roadside cameras for vehicle trajectory and speed profile estimation. Dynamic speed guidance algorithm is proposed to help on-ramp vehicles to merge into mainline smoothly, even in non-cooperative environment where mainline vehicles are not expected to slow down to accommodate on-ramp vehicles. On-site experiments were taken out in a merging area of Hangzhou Belt Highway to testify our prototype system, and simulation analysis shows our proposed algorithm can achieve significant fuel savings during the ramp merging process.

## 1 INTRODUCTION

Conflicts in highway ramp merging areas contribute significantly to traffic congestion, excessive fuel consumption, as well as traffic accidents on highways (Ahammed et al., 2008). Traditional approaches to address this issue are primarily to control the volume or the speed of the traffic on ramp and mainline (Milanes et al., 2011), and there are various applications such as ramp metering (Papageorgiou et al., 1990) variable speed limits (Ke et al., 2021), hard shoulder running (Li et al., 2014), and etc. However, limited by how instructions are notified to drivers, which is primarily through traffic signals, signs and roadside information board, the present control strategies are relatively static over time and undifferentiated towards different vehicles.

Along with the development of cooperative intelligent transportation system (C-ITS) and connected and autonomous vehicle (CAV) technologies, it helps to expand the research scope and the control measures of intelligent transportation system. Advanced wireless communication technology allows CAVs to report detailed vehicular data (e.g. speed, position, motion planning) and to receive abundant traffic data (e.g. high-definition map, personalized route planning suggestion, real-time traffic status) (Malik et al., 2021). Considering the ability to communicate among CAVs and the infrastructure, it draws attentions in academia to study the merging control strategies for CAVs at highway on-ramps (Zhu et al., 2022).

Single-lane highways with only CAVs is the most widely analyzed scenario, in which case it assumes that all vehicles are CAVs and there are no lane-changing maneuvers in the most-outer lane of mainline (Figure 1-a). To result in smooth on-ramp merging, it requires a merging sequence and a series of acceleration/deceleration instructions for all related vehicles. Merging sequence can be determined based on simple rule such as first-in-first-out, or to be optimized by considering the estimated arrival time (Xu et al., 2019). There are also studies integrated the merging sequence optimization with vehicle trajectory planning to simultaneously determine appropriate merging gap (Chen et al., 2021a). Given the merging sequence, a typical approach to generate the detailed acceleration/deceleration instruction is by mapping on-ramp vehicles to the mainline, which is also called "virtual vehicles", so as to convert the problem to a car-following/platooning control problem (Hu et al., 2021). There are also studies that build an optimization framework, with optimization objectives such as to maximize road capacity, minimize fuel consumption or maximize passenger's comfort level, and it results in desired position and speed of merging vehicles after controls (Zhou et al., 2019).

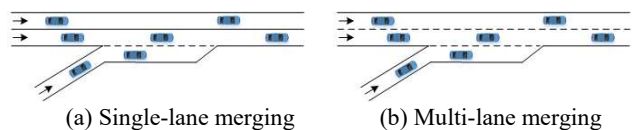

(a) Single-lane merging    (b) Multi-lane merging
Figure 1 On ramp merging scenarios

It is a broad consensus that there will be a long period of time that CAVs and human-drive vehicles (HDVs) share the roads before CAVs are fully penetrated on roads. As the presence of HDVs brings various uncertainties, mixed traffic situation is a great challenge for CAV control. A common approach is to build a model prediction control framework to predict HDVs trajectory and then to correspondingly adjust the speed of merging CAVs or the CAVs on mainline (Kherroubi et al., 2022). Some studies build a feedback control framework to further consider the deviation between the predicted and the observed HDVs' trajectory (Omidvar et al., 2020). What's more, there are some studies furtherly relaxed the assumption of no lane-changing maneuvers in the most-outer lane of mainline (Figure 1-b), and allowed flexible lane-changing choices for all vehicles in a two-lane highway (Williams et al., 2022).

However, considering the low penetration rate of CAVs at present, it is very unlikely that when a CAV on ramp wants to merge into the mainline, the neighboring vehicles on the mainline happens to be also CAVs. Thus, to design a feasible on-ramp merging dynamic speed guidance algorithm for the present, it has to obey the prevailing fact that vehicles on the highway mainline have the right-of-way over vehicles on ramps. In another word, vehicles on the highway mainline are not expected to slow down to accommodate vehicles on ramps.

Thus, it is more practical to assume that on-ramp merging area is a non-cooperative environment. This paper proposed a relative straightforward yet robust approach to generate dynamic speed guidance for CAV on-ramp merging in the single-lane highway with mixed traffic scenario. An on-site experiment was conducted to testify that our approach can effectively reduce the fleet energy consumption by avoiding hard acceleration/deceleration during the on-ramp merging process. In this experiment, a complete C-ITS prototype, which consists of hardware and software, including roadside camera-only perception, computing and communication, was implemented in an on-ramp area of Hangzhou Belt Highway, and the prototype system have been running for four months up-to-now to validate the feasibility of our total solution.

## 2 METHODOLOGY

As a complete C-ITS prototype, the general workflow of the dynamic speed guidance for on-ramp merging consists of three steps. Considering most on-road vehicles are ordinary vehicles without on-board communication unit (OBU) to report their vehicular data, so the first step is to identify vehicles and to estimate their positions from the raw images captured by the roadside camera. With a series of vehicle positions, the second step is to restore the vehicle trajectories and speed profile. Finally, the third step is to generate the dynamic speed guidance instructions based on the relative position relation of vehicles in the merging area.

### 2.1 Vehicle Tracking

An end-to-end vehicle tracking model was constructed, which consists of two deep neural networks (DNNs), as Figure 2Figure 1 shows.

The first DNN is designed for vehicle detection task (labeled as DNN1 in Figure 2), which consists of 2D BackBone, Detection Head, and post process layers. DNN1 generates target tensor $F(i)$ from the Detection Head layer, as well as target rectangular $R(i)$, category $C(i)$, and detection confidence $S(i)$ of each target $i$ in the input image.

The second DNN is designed for vehicle tracking task (labeled as DNN2 in Figure 2), which consists of Roi Align, 2D Conv. and FC layers. DNN2 takes the target tensor $F(i)$ and the target rectangle $R(i)$ from DNN1 as inputs, and it generates ID feature $IF(i)$ of each target $i$.

Targets are reidentified across frames based on the $IF$ using the cosine similarity and the Hungarian algorithm. Successfully reidentified targets will inherit their $ID$ from the previous frame, while new $ID$ will be assigned to the rest of targets that are failed to be reidentified.

Based on the aforementioned process, the model will eventually generate the target rectangle $R(i)$, category $C(i)$, detection confidence $S(i)$ and $ID(i)$ of each target $i$ in the newly captured video frame.

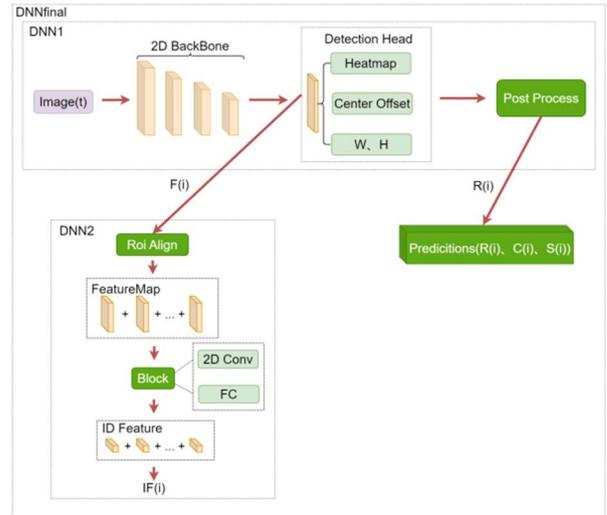

Figure 2 Vehicle tracking model structure

### 2.2 Trajectory Denoising

The position of vehicle $i$ is inferred as the center of the target rectangle $R(i)$ in each video frame. Thus, it is conceivable that even without misidentification, there are still errors in the vehicle position estimation, such as the camera shake caused by wind, the irregular oscillation of the vehicle position estimation caused by the tracking rectangle vibration, and the background interference. Without necessary intermediate denoising, it will cause cumulative errors in trajectory restore.

There are two types of denoising algorithms that are commonly used for the position estimation problem (Bruno et al., 2006): one is linear processing such as moving average method and exponential smoothing method, the other is non-linear processing such as wavelet analysis and Fourier analysis. Due to the randomness and instability of traffic flow changes, the obtained trajectory data is a non-stationary signal with random and irregular noise. According to the characteristics of this data signal, the wavelet transform method (Hobbs et al., 2014) which can realize high-resolution local analysis in both time and frequency domains is more suitable (Zhang et al., 2015).

The basic principle of wavelet transform is to suppress the useless part of signal and to enhance the useful part. A wavelet transform filter decomposes raw trajectory data $T^n$ into scaling and wavelet subsets under a given basis. The wavelet subsets contain details and noises in the raw trajectory. The irregular oscillation in the raw trajectory results in high fluctuation margin of wavelet subsets, which can be considered as white noises. We can suppress these noises by setting up appropriate thresholds, and obtain clean trajectory $T^c$ by combining the scaling and non-noise wavelet trajectory subsets.

The general denoising process is shown in Figure 3.

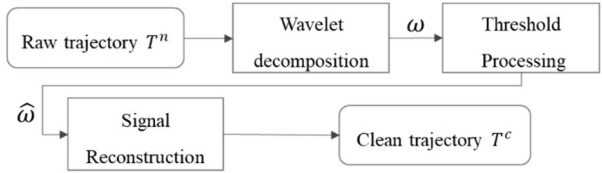

Figure 3 Denoising process

2.2.1 Wavelet decomposition

An appropriate wavelet basis function and a J-scale decomposition is applied to the $T^n$ to obtain the approximate-trajectory part $a_J = <T^n, \varphi_J(t)>$ and the detail trajectory parts $d_j = <T^n, \psi_j(t)>$ where $j = 1, ... J$. $\varphi_J(t)$ and $\psi_j(t)$ are the wavelet basis function and scale function, respectively.

To the author's knowledge, there is no mature theory to guide how to select the wavelet basis function. However, many wavelet analysis practices show that wavelet basis functions such as Haar and Daubechies often get better results (Chen et al., 2021b). In this study, the db3 basis from Daubechies family with a level number of 3 was selected to smooth the outliers in the original vehicle tracking data.

2.2.2 Threshold processing

After wavelet decomposition, each layer's coefficients are quantified by selecting a suitable threshold value, and then the quantized wavelet coefficients are obtained.

In terms of threshold setting, since the trajectory approximate factor $a_J$ is fixed, the trajectory detail factors $d_j(j = 1, ... J)$ are smoothed by wavelet decomposition. Specifically, considering $d_j(j = 1, ... J)$ have different signal-to-noise ratios (SNR), we set different thresholds for smoothing out noises in different trajectory detail signals. The rules of setting thresholds are given as follows:

a) As for the trajectory detail factor $d_J$, it contains most of the energy contributed to chaotic noises, and with higher SNR. Therefore, the threshold $t_J$ should not be too large to avoid removing the truly trajectory details.

$$t_J = \sigma\sqrt{2\ln N}/\sqrt{J} \quad (1)$$

b) As for the trajectory detail factor $d_j(j = 1, ... J - 1)$, there is a balance energy between chaotic signals and noise, and the threshold $t_j$ should be larger than tJ to successfully suppress noises in trajectory details, and the threshold is set as:

$$t_j = \sigma\sqrt{2\ln N}/\ln(j+1) \quad (2)$$

where $\sigma = \frac{median(|\omega|)}{0.6745}$, which is standard deviation of noise, $\omega$ is wavelet coefficient of signal, $N$ is the length of data, $j$ is the level number.

The threshold function is designed as:

$$\hat{\omega} = \begin{cases} \omega - at, & \omega \geq t \\ 0, & |\omega| < t \\ \omega + at, & \omega \leq t \end{cases} \quad (3)$$

where $\hat{\omega}$ is wavelet coefficients after threshold quantification, a is parameters for tuning parameters, $t$ is the thresholds identified above.

2.2.3 Signal reconstruction

By smoothing noises in the trajectory detail factor $d_j(j = 1, ... J - 1)$, it will eventually generate the denoised trajectory $T^c$ as:

$$T^c = a_J + \sum_{j=1}^{J} d_j' \quad (4)$$

where $d_j'$ is the smoothed trajectory details.

2.3 Dynamic Speed Guidance

Given the trajectories of vehicles in the ramp merging area, the speed suggestion for on-ramp merging can be calculated dynamically. The general idea is to map the leading vehicle on ramp to the most-outer lane of mainline as a virtual vehicle, then find an appropriate vehicle on the same lane as the leader of the virtual vehicle, and implement intelligent driver model (IDM) (Treiber et al., 2000) to perform car-following behavior.

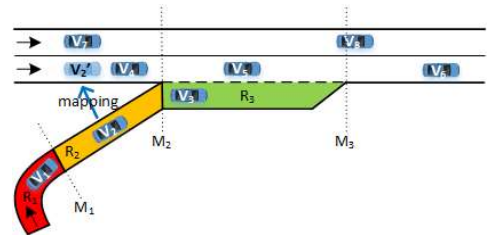

Figure 4 Ramp segmentation and vehicle mapping

As Figure 4 shows, an on-ramp can be divided into three segments. R1 is usually a big curvature segment which connects the local road, and the corresponding speed limit is relatively low. R2 is usually a relative straight and long segment, which allows vehicles to accelerate to approach the average speed of the mainline. R3 is the segment where the lane-changing behavior is executed.

For each time step, there are four key tasks:

1. Updating vehicle position: after the aforementioned vehicle tracking and trajectory denoising towards the roadside video, on-ramp vehicles can be divided into three groups according to their positions, including vehicles in R1, R2, and R3.

2. Execute lane changing behavior: for vehicles in the R3 segment (e.g. V3 in Figure 4), they are expected to be ready to execute lane changing behavior (including the available gap and the relative speed difference compared with the target lane) after the dynamic speed guidance process in R2. If the real-time traffic is not as expected, then the fourth step of update speed guidance will keep executing towards the ineligible-lane-changing vehicles in the R3 segment.

3. Search on-ramp target vehicle: if there are ineligible-lane-changing vehicles in the R3 segment, the leading vehicle in the R3 segment is the on-ramp target vehicle for the fourth step of speed guidance update. Otherwise, the leading vehicle in the R2 segment (e.g. V2 in Figure 4) will be chosen as the on-ramp target vehicle for the fourth step.

4. Update speed guidance: by mapping the on-ramp target vehicle to the most-outer lane of mainline as a virtual vehicle (e.g. V2' in Figure 4), searching the corresponding leader vehicle of the virtual vehicle (e.g. V4 in Figure 4), then the speed guidance can be calculated according to the position and speed of the ego (virtual on-ramp target) vehicle and the leading vehicle according to Equation (5)-(6).

$$a = a_m [1 - \left(\frac{v}{v_{max}}\right)^\delta - \left(\frac{s^*}{s}\right)^2] \quad (5)$$

$$s^* = s_{min} + T_s v + \frac{v \Delta v}{2\sqrt{a_m b_n}} \quad (6)$$

where $a$ is the recommended acceleration, $s^*$ is the desired gap, $s$ is the observed gap of between ego vehicle and its leading vehicle, $v$ is the speed of ego vehicle, $\Delta v$ is the speed difference between ego vehicle and its leading vehicle, $v_{max}$ is the speed limit, $s_{min}$ is the minimum vehicle spacing, $T_s$ is the desired time headway, $a_m$ is the maximum acceleration, $b_n$ is desired deceleration, $\delta$ is the free-drive exponent.

## 3 EXPERIMENT DESIGN

### 3.1 On-site Experiments

An on-ramp at the SanDun Entry of Hangzhou Belt Highway was chosen to implement a complete C-ITS prototype system, to demonstrate the function of dynamic speed guidance for CAV on-ramp merging, and to testify the performance of our proposed algorithm. The prototype system consists of a roadside camera and a roadside communication unit (RSU) at the top of roadside pole (Figure 5-b), and a small industrial computer at the bottom of the roadside pole (Figure 5-a).

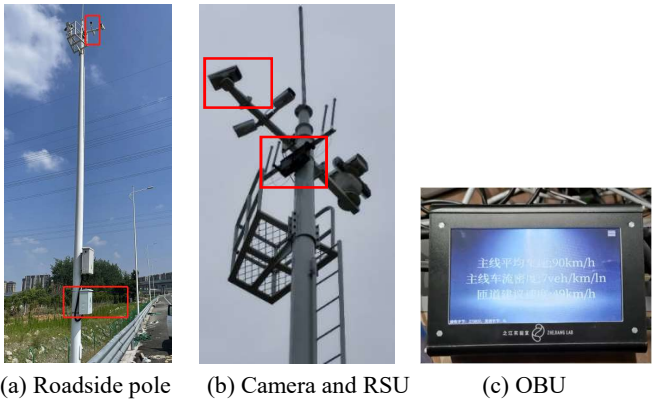

(a) Roadside pole    (b) Camera and RSU    (c) OBU

Figure 5 On-site experiment hardware

The roadside camera has a global view of the on-ramp merging area (Figure 6), and the live video streaming is processed to generate speed guidance instruction by every second. The effective communication distance of the RSU is over 350m, and it allows vehicles equipped with an OBU to receive the speed guidance instruction. For safety concern, the CAV is driven by human in this experiment, and the driver is asked to follow the speed guidance instruction displayed on the OBU screen (Figure 5-c).

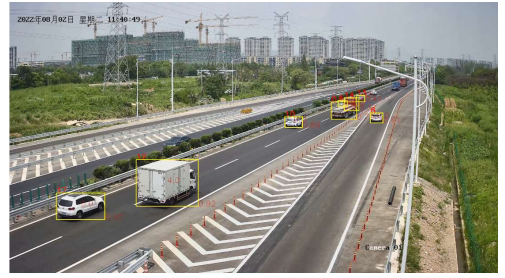

Figure 6 View of the roadside camera and vehicle tracking results

### 3.2 Simulation Experiments

Since traffic flow can hardly be replicated in a real road environment, it is difficult to testify the effectiveness of our proposed algorithm by controlling variates. Therefore, the observed traffic flow of the highway mainline is duplicated in the simulation environment for further experiments. As discussed in the introduction section, vehicles on the highway mainline have the right-of-way over vehicles on ramps, so fixed trajectories for vehicles on the mainline is consistent with our assumption that vehicles on mainline won't slow down to accommodate on-ramp vehicles.

In addition to the observed traffic flow, more scenarios were conducted in the simulation environment to test the performance, adaptability and robustness of our proposed algorithm in different scenarios. Examined variables include the traffic density of mainline and on-ramp, the length of on-ramp segment R2 and R3, the average start speed at the start point of R2, and whether the vehicles on mainline perform cooperative lane changing or not.

Simulation analysis is conducted using SUMO, simulated fuel consumption (SUMO, 2022a) is used for performance comparison, and the cooperative lane changing is simulated by defining the lane change mode (SUMO, 2022b) of vehicles on mainline. In the baseline scenario, the car-following model is set to be IDM, and the lane change mode is set to do strategic changes but no cooperative changes. All scenarios are repeated by 100 times with different random seeds to examine the probability distribution.

In the results and discussion section, the fuel saving is calculated as:

$$Fuel\ Saving = 1 - \frac{Fuel\ Consumption_{New}}{Fuel\ Consumption_{Baseline}} \quad (7)$$

## 4 RESULTS AND DISCUSSION

### 4.1 Trajectory Denoising

By comparing the original and the denoised vehicle trajectories, it can be clearly found that the vehicle trajectory became smoother (Figure 7) and the abnormal speed fluctuation became smaller (Figure 8) after the denoising process, which is more in line with our common sense expectations. Since in this experiment, the OBU is post-installed and stand alone, it is not able to retrieve vehicular information such as position or speed. We are still working on data interchange between the OBU and the vehicle, and it will help to provide true values of vehicular position and speed profile for the trajectory denoising performance analysis in our future work.

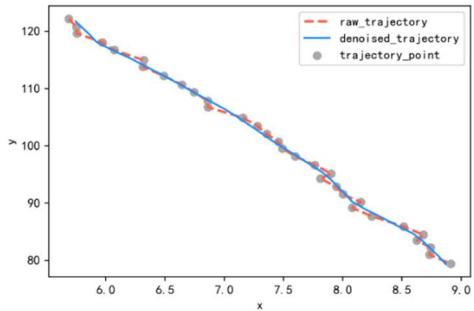

Figure 7 Original vs. denoised vehicle trajectory

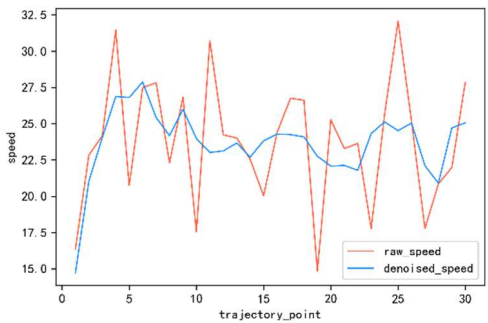

Figure 8 Original vs. denoised vehicle speed profile

*4.2 Dynamic Speed Guidance*

According to simulation analysis, the proposed dynamic speed guidance algorithm can achieve significant fuel savings. With a fixed mainline traffic density of 2000 veh/hr/ln, the fuel saving achieved by our proposed algorithm increases from 18% to nearly 90% along with the ramp traffic density increases in the non-cooperative case (Figure 9). Because it is assumed that vehicles on the mainline will not perform cooperative lane change and the car-following model is set to be IDM, it results in relative conservative lane-changing strategy. As a result, it causes traffic congestion on ramp when the ramp traffic density is 300-400 veh/hr/ln or more in the baseline scenario, which results in the dramatic increasement in fuel saving at the corresponding section. However, it is a fair comparison because in the contrasting scenario of dynamic speed guidance, the lane changing criteria also follows the IDM requirement.

Even in the cooperative lane-changing case, our proposed dynamic speed guidance algorithm can still achieve relative stable performance of 10% - 20% fuel savings (Figure 9). When the ramp traffic density exceeds 600 veh/hr/ln which might only happen during the rush hours, the fuel saving performance decreases along with traffic increasement.

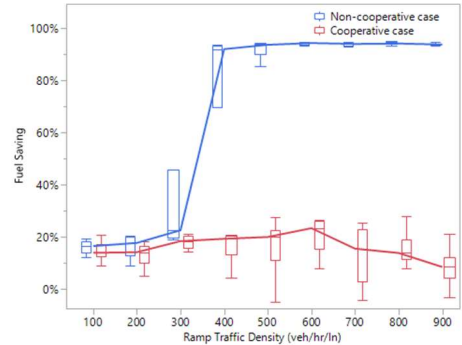

Figure 9 Fuel saving by ramp traffic density

Given a fixed ramp traffic density of 200 veh/hr/ln, the impact of mainline traffic density is further examined (Figure 10). With the mainline traffic density is relatively low, our proposed algorithm can achieve over 25% fuel saving. Once the mainline traffic density exceeds 1600 veh/hr/ln, the achievable fuel saving decreases along with mainline traffic density increasement. Especially when the mainline traffic density reaches 3000 veh/hr/ln, there is a chance that our algorithm performs even worse than without any dynamic speed guidance.

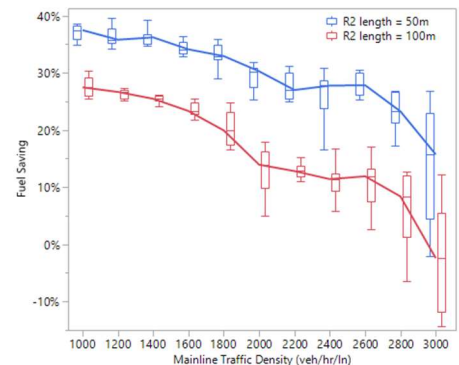

Figure 10 Fuel saving by mainline traffic density

We think there are primarily two factors. One is that IDM has relative conservative expectation for vehicle gap, and the other is when the length of acceleration segment (R2 in Figure 4) is too long, it increases the uncertainty of choosing the right appropriate leading vehicle to follow. Because on-ramp vehicles are mapped to the mainline once they enter the R2 segment. If a leading vehicle is chosen too early, it increases the chance of leading vehicle replacement, especially when the mainline traffic density is high. Therefore, we further shorten the length of R2 segment by half from 100 meters to 50 meters. As the Figure 10 shows, it significantly helps improve the performance of our algorithm in fuel saving by shortening the spatial extent of searching and adapting the leading vehicle on mainline for the on-ramp vehicles.

## 5 CONCLUSIONS

This paper presented a series of methods, including a DNN model for vehicle tracking from raw roadside camera data, a wavelet transform model for vehicle trajectory denoising, and then to generate dynamic speed guidance instruction for on-ramp vehicle merging. Particularly, our study considered a non-cooperative scenario that vehicles on mainline are ordinary HDVs without CAV communication capability.

Therefore, our proposed algorithm is feasible for the present mixed-traffic environment. A C-ITS prototype system was implemented in a freeway on-ramp merging area for testification purpose. Besides, simulation analysis shows that our proposed dynamic speed guidance algorithm can achieve significant fuel saving in both non-cooperative and cooperative scenarios.

ACKNOWLEDGEMENT

This work was supported by the Provincial Key R&D Program of Zhejiang (project number: 2021C01012).

The authors would like to thank Hangzhou Belt Highway for their supports to our on-site experiment.